\documentclass[runningheads]{llncs}
\usepackage{graphicx}
\usepackage{amsmath,amssymb} 
\usepackage{color}
\usepackage{csquotes}
\usepackage{cite}

\usepackage[small]{caption}
\usepackage{booktabs}       
\usepackage{mathtools}
\usepackage{xspace}

\begin{document}
\pagestyle{headings}
\mainmatter
\def\ECCV16SubNumber{***}  

\title{Category-level 6D Object Pose Recovery in Depth Images} 



\author{Caner Sahin, Tae-Kyun Kim}
\institute{ICVL, Imperial College London}

\maketitle

\begin{abstract}
Intra-class variations, distribution shifts among source and target domains are the major challenges of category-level tasks. In this study, we address category-level full 6D object pose estimation in the context of depth modality, introducing a novel part-based architecture that can tackle the above-mentioned challenges. Our architecture particularly adapts the distribution shifts arising from shape discrepancies, and naturally removes the variations of texture, illumination, pose, etc., so we call it as \enquote{Intrinsic Structure Adaptor (ISA)}. We engineer ISA based on the followings: i) \enquote{Semantically Selected Centers (SSC)} are proposed in order to define the \enquote{6D pose} at the level of categories. ii) 3D skeleton structures, which we derive as shape-invariant features, are used to represent the parts extracted from the instances of given categories, and privileged one-class learning is employed based on these parts. iii) Graph matching is performed during training in such a way that the adaptation/generalization capability of the proposed architecture is improved across unseen instances. Experiments validate the promising performance of the proposed architecture using both synthetic and real datasets.
\end{abstract}
\vspace{-2.4em}
\section{Introduction}
\label{Introduction}
Accurate 3D object detection and pose estimation, also known as 6D object pose recovery, is an essential ingredient for many practical applications related to scene understanding, augmented reality, control and navigation of robotics, \textit{etc}. While substantial progress has been made in the last decade, either using depth information from RGB-D sensors \cite{18, 31, 36, 77, 76, 71, 74, 73} or even estimating pose from a single RGB image \cite{21, 68, 69, 70}, improved results have been reported for instance-level recognition where source data from which a classifier is learnt share the same statistical distributions with the target data on which the classifiers will be tested. Instance-based methods cannot easily be generalized for category-level tasks, which inherently involve the challenges such as distribution shift among source and target domains, high intra-class variations, and shape discrepancies between objects, \textit{etc}.\\
\indent At the level of categories, Sliding Shapes (SS) \cite{2}, an SVM-based method enlarging search space to 3D, detects objects in the context of depth modality naturally tackling the variations of texture, illumination, and viewpoint. The detection performance of this method is further improved in Deep Sliding Shapes (Deep SS) \cite{1}, where more powerful representations encoding geometric shapes are learned in ConvNets. These two methods run sliding windows in the 3D space mainly concerning 3D object detection rather than full 6D pose estimation. The system in \cite{4}, inspired by \cite{2}, further estimates detected and segmented objects' rotation around the gravity axis using a CNN. The system is the combination of individual detection/segmentation and pose estimation frameworks. Unlike these methods, we aim to directly hypothesise full 6D poses in a single-shot operation. The ways the methods above \cite{2, 1, 4} address the challenges of categories are relatively naive. Both SS and the method in \cite{4} rely on the availability of large scale 3D models in order to cover the shape variance of objects in the real world. Deep SS performs slightly better against the categories' challenges, however, its effort is limited to the capability of ConvNets.\\
\begin{figure}[!t]
\captionsetup[subfigure]{labelformat=empty}
\centering
\includegraphics[height=1.55in]{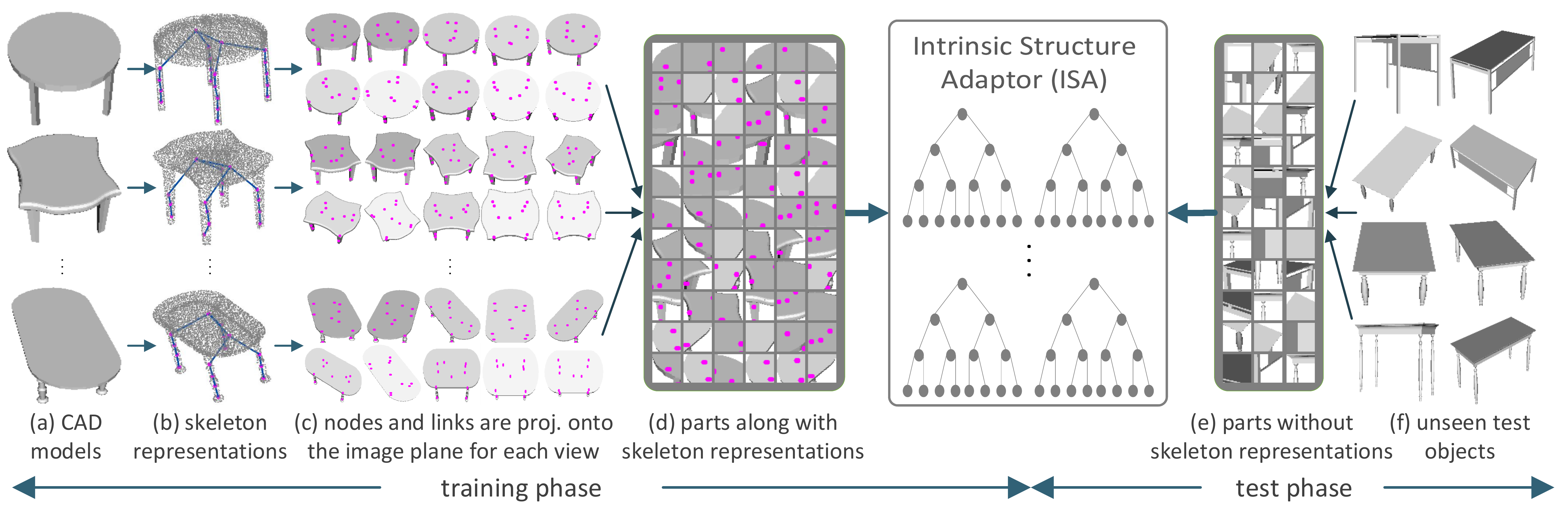}
\caption{ Intrinsic Structure Adaptor (ISA) is trained based on parts extracted from instances of a given category. CAD models in (a) are represented with skeletons in (b). Nodes and links are projected onto the image plane in (c) for each view. Parts along with skeletal representations in (d) are fed into the forest. In the test, appearances of the parts (e) that are extracted from depth images of unseen instances (f) are used in order to hypothesise 6D pose.}
\label{figminus1}
\vspace{-1.0em}
\end{figure}
\indent In this study, we engineer a dedicated architecture that directly tackles the challenges of categories while estimating objects' 6D. To this end, we utilize \textit{3D skeleton structures}, derive those as shape-invariant features, and use those as privileged information during the training phase of our architecture. \textit{3D skeleton structures} are frequently used in the literature in order to handle shape discrepancies \cite{60, 61, 15, 16}. We introduce \enquote{Intrinsic Structure Adaptor (ISA)}, a part-based random forest architecture, for full 6D object pose estimation at the level of categories in depth images. ISA works in the 6D space. It neither requires a segmented/cropped image as in \cite{4}, nor asks for 3D bounding box proposals as in \cite{1}. Unlike \cite{2, 1}, instead of running sliding windows, ISA extracts parts from the input depth image, and feeding all those down the forest, directly votes for the 6D pose of objects. Its training phase is processed so that the challenges of the categories can successfully be tackled. 3D skeleton structures are used to represent the parts extracted from the instances of given categories, and privileged learning is employed based on these parts. Graph matching is performed during the splitting processes of random forest in such a way that the adaptation/generalization capability of the proposed architecture is improved across unseen instances. Note that, unlike \cite{2, 1, 4}, this is one-class learning, and a single classifier is learnt for all instances of the given category. Figure \ref{figminus1} depicts the whole system of our architecture. To summarize, our main contributions are as follows:\\
\noindent \textbf{Contributions.} \enquote{Semantically Selected Centers (SSC)} are proposed in order to define the \enquote{6D pose} at the level of categories. 3D skeleton structures, which we derive as shape-invariant features, are used to represent the parts extracted from the instances of given categories, and privileged one-class learning is employed based on these parts. Graph matching is performed during training in such a way that the adaptation/generalization capability of the proposed architecture is improved across unseen instances.
\vspace{-1.0em}
\section{Related Work}
\label{Related_Work}
A number of methods have been proposed for 3D object detection and pose estimation, and for skeleton representations. For the reader's convenience, we only review 6D case for instance-level object detection and pose estimation, and keep category-level detection broader.
\vspace{-1em}
\subsection{Object Detection and Pose Estimation}
\textbf{Instance-level (6D):} State-of-the-art methods for instance-level 6D object pose estimation report improved results tackling the problem's main challenges, such as occlusion and clutter, and texture-less objects, \textit{etc}. The holistic template matching approach, Linemod \cite{32}, estimates cluttered object's 6D pose using color gradients and surface normals. It is improved by discriminative learning in \cite{34}, and later been utilized in a part-based random forest method \cite{20} in order to provide robustness across occlusion. Occlusion aware features \cite{30} are further formulated, and more recently feature representations are learnt in an unsupervised fashion using deep convolutional networks \cite{18, 33}. The studies in \cite{31, 36} cope with texture-less objects. Whilst these methods fuse data coming from RGB and depth channels, a local belief propagation based approach \cite{35} and an iterative refinement architecture \cite{17, 19} are proposed in depth modality \cite{75}. 6D pose estimation is recently achieved from RGB only \cite{21, 68, 69, 70}. Despite being successful, instance-based methods cannot easily be generalized for category-level tasks, which inherently involve the challenges such as distribution shift among source and target domains, high intra-class variations, and shape discrepancies between objects, \textit{etc}.\\
\textbf{Category-level:} At the level of categories, several studies combine depth data with RGB. Depth images are encoded into a series of channels in \cite{28} in such a way that R-CNN, the network pre-designed for RGB images, can represent that encoding properly. The learnt representation along with the features extracted from RGB images are then fed into an SVM classifier. In another study \cite{27}, annotated depth data, available for a subset of categories in Imagenet, are used to learn mid-level representations that can be fused with mid-level RGB representations. Although promising, they are not capable enough for the applications beyond 2D.\\
\indent Sliding Shapes (SS) \cite{2}, an SVM-based method, hypothesises 3D bounding boxes of the objects, and naturally tackles the variations of texture, illumination, and viewpoint, since it works in depth images. However, hand-crafted features used by the method, being unable to reasonably handle the challenges of categories, limit the method's detection performance across unseen instances. Deep Sliding Shapes (Deep SS) \cite{1}, the method based on 3D convolutional neural networks (CNN), learns more powerful representations for encoding geometric shapes further improving SS. However, the improvement is architecture-wise, and Deep SS encodes a 3D space using Truncated Signed Distance Functions (TSDF), similar to SS. Although promising, both methods concentrate on hypothesising 3D bounding boxes, running sliding windows in the 3D space. Our architecture, ISA, works in the 6D space. Instead of running sliding windows, it directly votes for the 6D pose of the objects by passing the parts extracted from the input depth image down all the trees in the forest. The system in \cite{4}, inspired by \cite{2}, further estimates detected and segmented objects’ rotation around gravity direction using a CNN, which is trained using pixel surface normals. A relative improvement is observed in terms of accuracy, however, the system is built integrating individual detection/segmentation and pose estimation frameworks. ISA neither requires a segmented/cropped image as in \cite{4}, nor asks for 3D bounding box proposals as in \cite{1}.\\
\indent Despite being proposed to work in large-scale scenarios, the methods \cite{2, 1, 4} do not have specific designs that can explicitly tackle the challenges of categories. SS relies on the availability of large scale 3D models in order to handle distribution shifts arising from shape discrepancies. Deep SS learns powerful 3D features from the data via a CNN architecture, however, the representation used to encode a 3D space is similar to the one used in SS, that is, the improvement on the feature representation arises from the CNN architecture. Gupta et al. \cite{4} use objects' CAD models at different scales in order to cover the shape variance of the objects in the real world while estimating objects' rotation and translation. Unlike these methods, ISA is a dedicated architecture that directly tackles the challenges of the categories while estimating objects’ 6D. It employs graph matching during forest training based on the parts represented with skeleton structures in such a way that the adaptation/generalization capability is improved across unseen instances.
\vspace{-1em}
\subsection{Skeleton Representation}
Skeletal structures have frequently been used in the literature, particularly to improve the performance of action/activity recognition algorithms. Baek et al. \cite{62} consider the geometry between scene layouts and human skeletons and propose kinematic-layout random forests. Another study \cite{61} utilizes skeleton joints as privileged information along with raw depth maps in an RNN framework in order to recognise actions. The study in \cite{15} shows that, one can effectively utilize 3D skeleton structures for overcoming intra-class variations, and for building a more accurate classifier, advocating the idea, domain invariant features increase generalization, stated in \cite{16}.
\vspace{-1.4em}
\section{Proposed Architecture}
\label{Proposed_Architecture}
This section presents the technologies top of which the proposed architecture, ISA, is based on. We firstly define the \enquote{pose} for the category-level 6D object pose estimation problem, and demonstrate the dataset and annotations discussing shape-invariant feature representations. We next present privileged one-class learning where we employ graph matching, and lastly we describe the test step, category-level 6D object pose estimation.
\vspace{-1em}
\subsection{Pose Definition: Semantically Selected Centers (SSC)}
\label{Pose_Definition}
A method designed for 6D pose estimation outputs the 3D position and 3D rotation of an object of interest in camera-centered coordinates. According to this output, it is important to precisely assign the reference coordinate frame to the interested object. When the method is proposed for instance-level 6D object pose estimation tasks, the most common approach is to assign the reference coordinate frame to the center of mass (COM) of the object's model. At the level of instances, source data from which a classifier is learnt share the same statistical distributions with the target data on which the classifiers will be tested, that is, training and test samples are of the same object. Hence, instance-level 6D pose estimators output the relative orientation between the COM of the object and the camera center. At the level of categories, in turn, this 6D pose definition cannot be directly utilised, since significant distribution shifts arise between training and test data.\\
\indent An architecture engineered for the category-level 6D object pose estimation problem should hypothesise 6D pose parameters of unseen objects. Objects from the same category typically have similar physical sizes \cite{1}. However, investigations over 3D models of the instances demonstrate that each instance has different COM, thus making the utilization of conventional 6D pose definition to malfunction for category-level tasks. In such a case, we reveal Semantically Selected Centers (SSC), which allow us to redefine the \enquote{6D pose} for the category-level 6D object pose estimation problem. For every category we define only one SSC performing the following procedure: 
\begin{figure}[!t]
\captionsetup[subfigure]{labelformat=empty}
\centering
\includegraphics[height=2.7in]{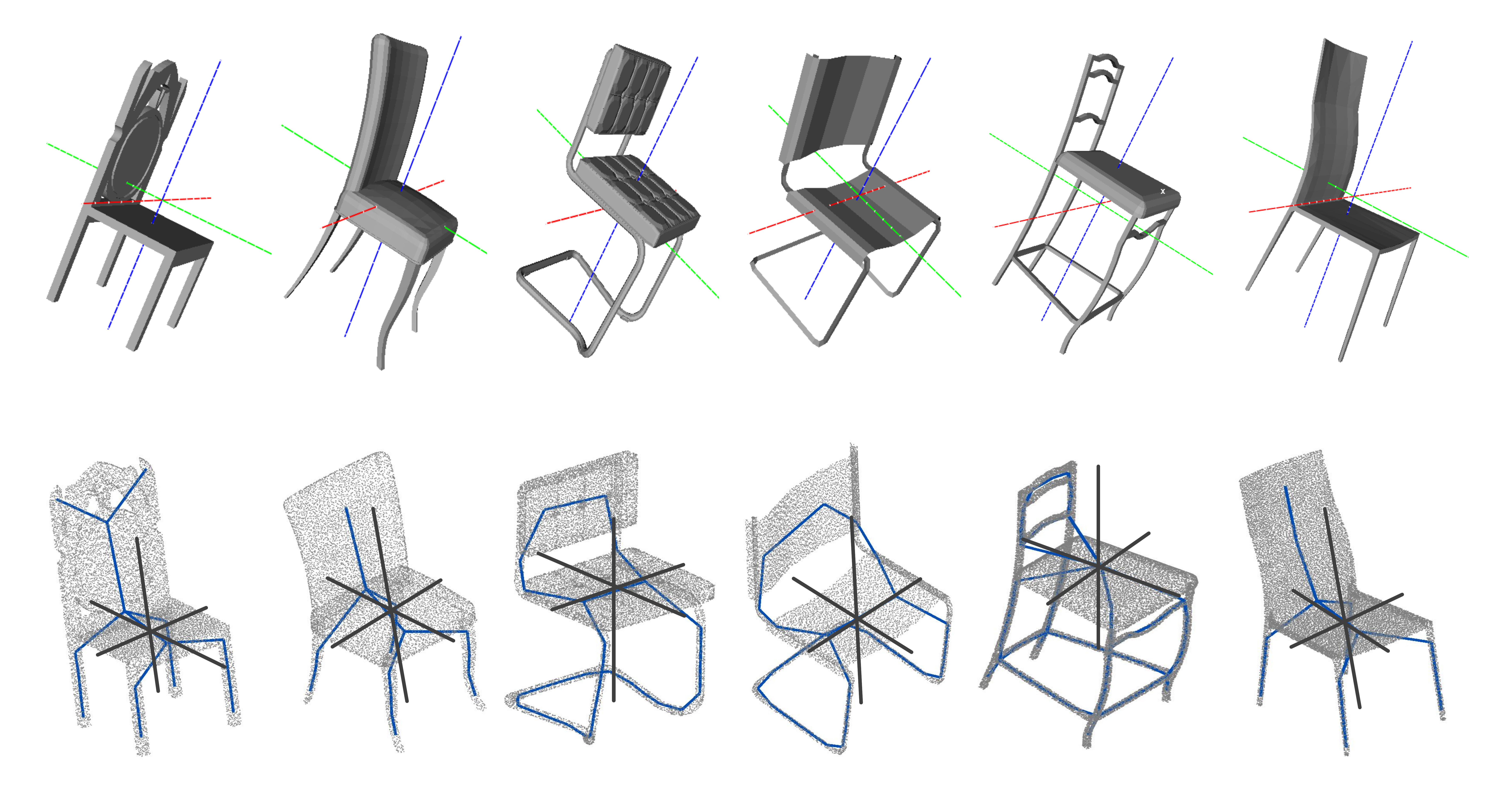}
\vspace{-2em}
\caption{Semantically Selected Centers (SSC): top row shows centers of mass of the instances, while the bottom row depicts SSCs of the corresponding instances (views best describing the difference selected).}
\vspace{-2em}
\label{fig1}
\end{figure}
\begin{itemize}
\item For each instance, skeletal graph representation is extracted, and the COM is found over 3D model. 3D distances between the nodes of the representation and the COM is computed.
\item Between all instances, the skeleton nodes are topologically matched, and the most repetitive node is determined.
\item In case there are more than $1$ repetitive node computed, the SSCs are determined by interpolating between the repetitive nodes.
\end{itemize}
Note that, this repetitive node is also the one closest to the COMs of the instances. As the last step, we assign reference coordinate frames to the related parts of the objects given in the category. Figure \ref{fig1} shows SSCs for the chair category. Despite the fact that COMs of the models are individually different, the 6D pose of each chair is defined with respect to Semantically Selected Centers (bottom row of the figure).\\
\indent The metric proposed in \cite{32}, Average Distance (AD), is designed to measure the performance of instance-level object detectors. In order to evaluate our architecture, we modify AD making this metric work at the level of categories via the Semantically Selected Centers (SSC) of the instances of the given category. $M^{{SSC}_i}_c$ is the 3D model of the instance $i$ that belongs to the category $c$, and the set of $M^{{SSC}_i}_c$ of the test instances form $\mathcal{M}_c$: $\mathcal{M}_c = \{ M^{{SSC}_i}_c | i = 1, 2, ... \}$. $X^{{SSC}_i}_c$ is the point cloud of the model $M^{{SSC}_i}_c$. Having the ground truth rotation $R$ and translation $T$, and the estimated rotation $\tilde{R}$ and translation $\tilde{T}$, we compute the average distance over $X^{{SSC}_i}_c$:
\vspace{-0.5em}
\begin{equation}
\omega_i = avg||(R X^{{SSC}_i}_c + T) - (\tilde{R} X^{{SSC}_i}_c + \tilde{T})||.
  \label{equation14}
\end{equation}
$\omega_i$ calculates the distance between the ground truth and estimated poses of the test instance $i$. The detection hypothesis that ensures the following inequality is considered as correct:
\begin{equation}
\omega_i \leq z_{\omega_i} \Phi_i
  \label{equation15}
\end{equation}
where $\Phi_i$ is the diameter of $M^{{SSC}_i}_c$, and $z_{\omega_i}$ is a constant that determines the coarseness of an hypothesis that is assigned as correct.
\begin{figure}[!t]
\captionsetup[subfigure]{labelformat=empty}
\centering
\includegraphics[height=1.1in]{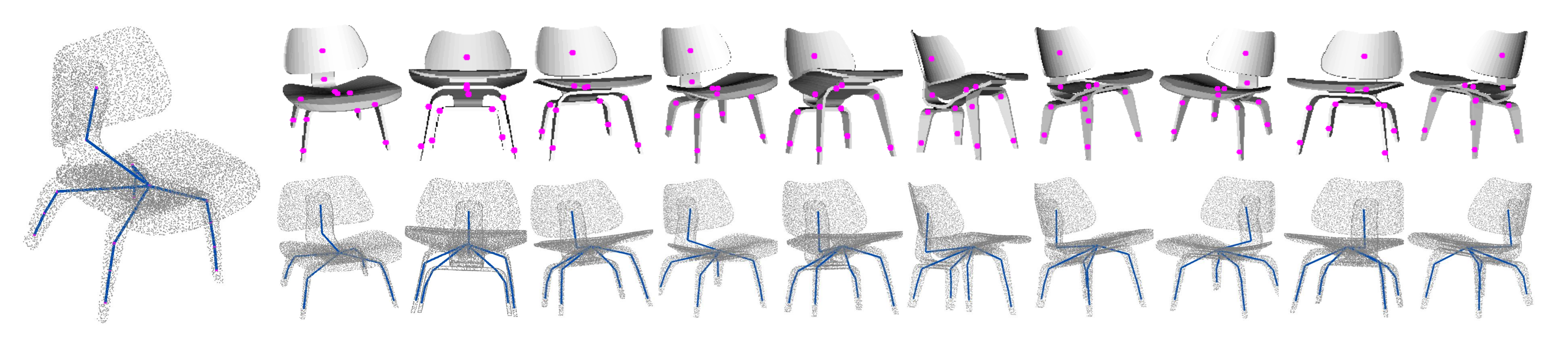}
\caption{Skeletal graph representation: skeleton nodes are determined with respect to model coordinate frame. Skeletal nodes and links are projected onto the image plane for each viewpoint at which a synthetic depth image is rendered.}
\vspace{-1.4em}
\label{fig2}
\end{figure}
\vspace{-1em}
\subsection{Dataset and Part Representations} 
The training dataset $\mathcal{S}$ involves synthetic data that are of $c_s$ instances of a given category. Using the 3D CAD models of these $c_s$ instances, we render foreground synthetic depth maps from different viewpoints and generate annotated parts in order to form $\mathcal{S}$:
\begin{equation}
\small
\begin{gathered}
\mathcal{S} = \{ \mathcal{P}_i \vert i = 1, 2, ..., c_s \}\\
\mathcal{P}_i = \{ \cup_{j=1}^{n} P_j \} = \{ \cup_{j=1}^{n} (\mathbf{c}_j, \Delta \mathbf{x}_j, \mathbf{\theta}_j, \mathbf{a}_j, \mathbf{s}_j, D_{P_j}) \}
\end{gathered}
\label{equation1}
\end{equation}
where $\mathcal{P}_i$ involves the set of parts $\{ P_j | j = 1, 2, ..., n\}$ that are extracted from the synthetic images of the object instance $i$. $\mathbf{c}_j = (c_{x_j}, c_{y_j}, c_{z_j})$ is the part centre in $[px,px,m]$. $ \Delta \mathbf{x}_j = (\Delta x, \Delta y, \Delta z)$ presents the 3D offset between the centre of the part and the SSC of the object, and $\theta_j = (\theta_r, \theta_p, \theta_y)$ depicts the 3D rotation parameters of the point cloud from which the part $P_j$ is extracted. $\mathbf{a}_j$ describes the vector of the skeletal link angles. $\mathbf{s}_j$ is the skeletal node offset matrix representation, and $D_{P_j}$ is the depth map of the part $P_j$.\\
\indent We next briefly mention how we derive $\mathbf{a}_j$ and $\mathbf{s}_j$ based on skeletal graph representation extracted from 3D model of an instance.\\
\noindent \textbf{Derivation of $\mathbf{a}_j$ and $\mathbf{s}_j$.} The algorithm in \cite{53} is utilized in order to extract the skeletal graph of an instance from its 3D model. Once the skeletal graph is extracted, we next project both the nodes and the links onto the image plane for every viewpoint at which synthetic depth maps are rendered. At each viewpoint, we measure the angles that the links of the graph representation make with the $x$ direction, and stack them into the vector of skeletal link angles $\mathbf{a}_j$ (see Fig. \ref{fig2}). All of the parts extracted at a specific viewpoint have the same representation $\mathbf{a}$. The distances between the centre $\mathbf{c}_j$ of each part $P_j$ and skeleton nodes are measured in image pixels along $x$ and $y$, and in metric coordinates along $z$ direction in order to derive the skeletal node offset matrix $\mathbf{s}_j$:
\begin{equation}
\mathbf{s}_j = [\Delta x_{j_i}, \Delta y_{j_i}, \Delta z_{j_i}]_{s_n \times 3}, \quad i = 1, 2, ..., s_n.
\label{equation6}
\end{equation}
\noindent Figure \ref{fig2} shows an example skeletal graph and its projection onto 2D image plane for several viewpoints. In this representation, we compute $19$ nodes in total and project onto the image plane $11$ of those.\\
\indent We next discuss how to handle shape discrepancies between the parts extracted from the instances of a given category using the representations $\mathbf{a}$ and $\mathbf{s}$.
\begin{figure}[!t]
\captionsetup[subfigure]{labelformat=empty}
\centering
\includegraphics[height=1.5in]{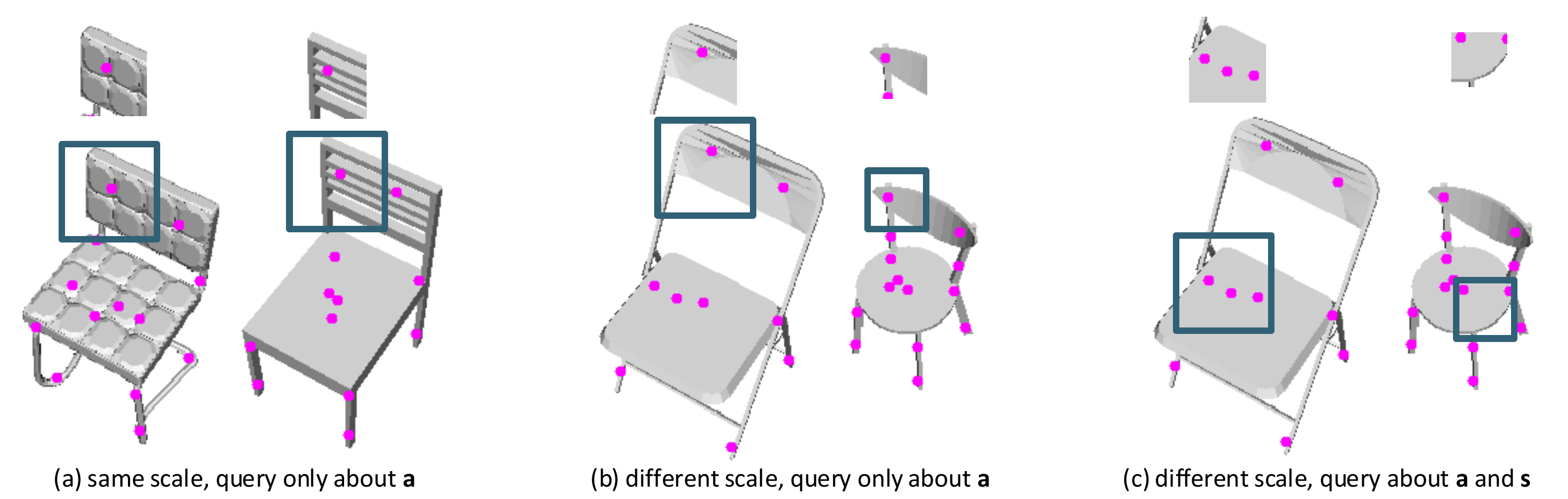}
\caption{Parts in (a) and (b) are topologically at the same location, having the same $\mathbf{a}$. Parts in (c) are topologically at different location, having the same $\mathbf{a}$, the case which is undesired. Hence the parts are further questioned with $\mathbf{s}$, the representation that removes mismatches.}
\vspace{-1.4em}
\label{fig3}
\end{figure}
\noindent \textbf{Privileged data: Shape-invariant skeleton representations.} We start our discussion by firstly representing the parts with $\mathbf{a}$. The study in \cite{1} states that objects from the same category typically have similar physical size, however, the appearances of the objects are relatively different. Figure \ref{fig3} (a) depicts the parts extracted from $2$ different objects, belonging to the same category. Despite the fact that both parts have different shapes in depth channel, their representations $\mathbf{a}$ are the same, tackling the discrepancy in shape.\\
\indent There are also cases where some instances are relatively larger in the given category. The vector of skeletal link angles, $\mathbf{a}$, readily handles the scale variation between the instances. In Fig. \ref{fig3} (b), the objects from which the parts extracted are different in both shape and in scale, however, the parts have the same representations $\mathbf{a}$. One drawback of this representation is that it is not sufficient enough to match topologically correct parts. In Fig. \ref{fig3} (c), the parts are semantically at different locations of the objects, however, they have the same $\mathbf{a}$. Hence, we additionally represent the parts with the skeletal node offset matrix $\mathbf{s}$. $\mathbf{s}$ along with $\mathbf{a}$ are used to adapt the intrinsic structures of the instances while topologically constraining the structures. In Fig. \ref{fig3} (c), when we query about $\mathbf{s}$, in addition to $\mathbf{a}$, the mismatch between the parts disappears, since both parts have different skeletal node offset matrix representation $\mathbf{s}$.\\
\vspace{-1.6em}
\subsection{Privileged One-Class Learning}
\label{Problem_Formulation}
ISA, being a part-based random forest architecture, is the combination of randomized binary decision trees. Employing one-class learning, it is trained only on positive samples, rather than explicitly collecting representative negative samples. The learning scheme is additionally privileged. The part representations $\mathbf{a}$ and $\mathbf{s}$ are only available during training, and not required during testing. This is achieved by using them in the split criteria (Eq. \ref{eq8_can}), but not in the split function (Eq. \ref{eq5_can}).  We use the dataset $\mathcal{S}$ in order to train ISA employing simple depth comparison features (2-pixel test) in the split nodes. At a given pixel $\mathbf{w}$, the features compute:
\begin{equation}
\small
f_\psi(D_P, \mathbf{w}) = D_P(\mathbf{w} + \frac{\mathbf{u}}{D_P(\mathbf{w})}) - D_P(\mathbf{w} + \frac{\mathbf{v}}{D_P(\mathbf{w})})
\label{eq5_can}
\end{equation}
where $D_P(\mathbf{w})$ is the depth value of the pixel $\mathbf{w}$ in part $P$, and the parameters $\psi = (\mathbf{u}, \mathbf{v})$ depict offsets $\mathbf{u}$ and $\mathbf{v}$. Each tree is constructed by using a randomly selected subset $\mathcal{W} = \{ P_j \}$ of the annotated training parts $\mathcal{W} \subset \mathcal{S}$. Starting from the root node, a group of splitting candidates $\{ \phi = (\psi, \tau) \}$, where $\psi$ is the feature parameter and $\tau$ is the threshold, are randomly produced. The subset $\mathcal{W}$ is partitioned into left $\mathcal{W}_l$ and right $\mathcal{W}_r$ by each $\phi$:
\begin{equation}
\small
\begin{gathered}
\mathcal{W}_l (\phi) = \{ \mathcal{W} | f_\theta(D_P, \mathbf{w}) < \tau \}\\
\mathcal{W}_r (\phi) = \mathcal{W} \setminus \mathcal{W}_l.
\end{gathered}
\end{equation}
The $\phi$ that best optimizes the following entropy is determined:
\begin{equation}
\small
\begin{gathered}
\phi^* = \arg \max_{\phi} (Q(\phi)) \\
Q = Q_1 + Q_2 + Q_3
\end{gathered}
\label{eq8_can}
\end{equation}
where $Q_1$, $Q_2$, and $Q_3$ are the 6D pose entropy, the skeletal link angle entropy, and the skeletal node offset entropy, respectively. Each tree is grown by repeating this process recursively until the forest termination criteria are satisfied. When the termination conditions are met, the leaf nodes are formed and they store votes for both the object center $\Delta \mathbf{x} = (\Delta x, \Delta y, \Delta z)$ and the object rotation $\theta = (\theta_r, \theta_p, \theta_y)$.\\
\noindent \textbf{Matching Skeletal Graphs.} When we build ISA, our main target is to provide adaptation between the instances, and to improve the generalization across unseen objects. Apart from the data used to train the forest, the quality functions we introduce play an important role for these purposes. The quality function $Q_1$, optimizing data with respect to only 6D pose parameters, is given below:
\begin{equation}
\small
Q_1 = log(|\Sigma^{\Delta \mathbf{x}}| + |\Sigma^{\theta}|) - \sum_{i \in {(L, R)}} {\frac{\mathcal{S}_i}{\mathcal{S}}} log(|\Sigma^{\Delta \mathbf{x}}_i| + |\Sigma^{\theta}_i|)
  \label{equation9}
\end{equation}
where $|\Sigma^{\Delta \mathbf{x}}|$, $|\Sigma^{\theta}|$ show the determinants of offset and pose covariance matrices, respectively. $\mathcal{S}_i$ depicts the synthetic data sent either to the left $L$ or to the right $R$ child node. In case the architecture is trained only using parts extracted from $1$ instance, $Q_1$ successfully works. We train ISA using multiple instances, targeting to improve the adaptation/generalization capability across unseen instances. In order to achieve that, we propose the following quality function in addition to $Q_1$:
\begin{equation}
\small
Q_2 = log(|\Sigma^{\mathbf{a}}|) - \sum_{i \in {(L, R)}} {\frac{\mathcal{S}_i}{\mathcal{S}}} log(|\Sigma^{\mathbf{a}}_i|)
  \label{equation9}
\end{equation}
where $|\Sigma^{\mathbf{a}}|$ shows the determinant of the skeletal link angle covariance matrix. This function measures the similarity of the parts regarding the angles that the links of the skeleton representations make with the $x$ direction. The main reason why we use this function is to handle shape discrepancies in depth channel between parts, even if the parts are extracted from relatively large scale objects. Let's suppose that if all parts under query are extracted from topologically same locations of the instances, the combination of $Q_1$ and $Q_2$ would be sufficient. On the other hand, the combination of these two functions is not sufficient, since the parts are extracted from the complete structures of the instances. In such a scenario, the parts coming from topologically different locations, but with similar $\mathbf{a}$ are tend to travel to the same child node, if the features used in the split function fails to correctly separate the data. Hence, we require the following function that prevents this drawback:
\begin{equation}
\small
Q_3 = log(|\Sigma^{\mathbf{s}}|) - \sum_{i \in {(L, R)}} {\frac{\mathcal{S}_i}{\mathcal{S}}} log(|\Sigma^{\mathbf{s}}_i|) \\
  \label{equation9}
\end{equation}
where $|\Sigma^{\mathbf{s}}|$ shows the determinant of the skeletal node offset covariance matrix. The main reason why we use $Q_3$ is to prevent topologic mismatches in between the parts extracted from different instances of the given category.
\vspace{-1em}
\subsection{Category-level 6D Object Pose Estimation}
Given a category of interest $c$, and a depth image $I^t$ in which an unseen instance of the interested category exists, the proposed architecture, ISA, targets to maximize the joint posterior density of the object position $\Delta \mathbf{x}$ and the rotation $\theta$:
\begin{equation}
\small
(\Delta \mathbf{x}, \theta) = \arg \max_{\Delta \mathbf{x}, \theta} p(\Delta \mathbf{x}, \theta \vert I^t, c).
\label{eq8}
\end{equation}
Since ISA is based on parts, and the parts extracted from $I^t$ are passed down all the trees by the split function in Eq. \ref{eq5_can}, we can calculate the probability $p(\Delta \mathbf{x}, \theta | I^t, c)$ for a single tree $T$ aggregating the conditional probabilities $p(\Delta \mathbf{x}, \theta | P, c)$ for each part $P$:
\begin{equation}
\small
p(\Delta \mathbf{x}, \theta \vert I^t, c; T) = \sum_i p(\Delta \mathbf{x}, \theta \vert P_i, c, D_{P_i}; T).
\label{equation12}
\end{equation}
In order to hypothesise the final pose parameters, we average the probabilities over all trees using the information stored in the leaf nodes for a given forest $F$:
\begin{equation}
\small
p(\Delta \mathbf{x}, \theta \vert I^t, c; F) = \frac{1}{\vert F \vert} \sum_t^{\vert F \vert} \sum_i p(\Delta \mathbf{x}, \theta \vert P_i, c, D_{P_i}; T_t).
  \label{equation13}
\end{equation}
\indent Please note that the above pose inference is done using a single depth image, not skeletons and their representations. 
\vspace{-1.0em}
\section{Experiments}
\label{Experiments}
\vspace{-0.5em}
In order to validate the performance of the proposed architecture, we conduct experiments on both synthetic and real data.\\
\noindent \textbf{Synthetic Dataset.} Princeton ModelNet10 dataset \cite{63} contains CAD models of $10$ categories, and in each category, the models are divided into train and test. We use the CAD models of the test instances of four categories, \textit{bed}, \textit{chair}, \textit{table}, and \textit{toilet}, and render depth images from different viewpoints, each of which is 6D annotated and occlusion/clutter-free. Each category involves $264$ images of unseen objects, and there are $1320$ test images in total. We compare ISA and instance-based Linemod on the synthetic dataset.\\
\noindent \textbf{Real Dataset.} RMRC \cite{64}, involving cluttered real depth images of several object categories, is the dataset on which we test and compare our architecture with the state-of-the-art methods \cite{2, 1, 4}. The images in this dataset are annotated only with 3D bounding boxes.\\
\noindent \textbf{Evaluation Protocols.} The evaluation protocol used for the experiments conducted on the synthetic dataset is the one proposed in Subsect. \ref{Pose_Definition}. We make use of the evaluation metric in \cite{2} when we compare ISA with the state-of-the-art methods on real data.\\
\vspace{-2.5em}
\subsection{Experiments on Synthetic Data}
\begin{figure*}[!t]
\captionsetup[subfigure]{labelformat=empty}
\centering
\includegraphics[height=2.8in]{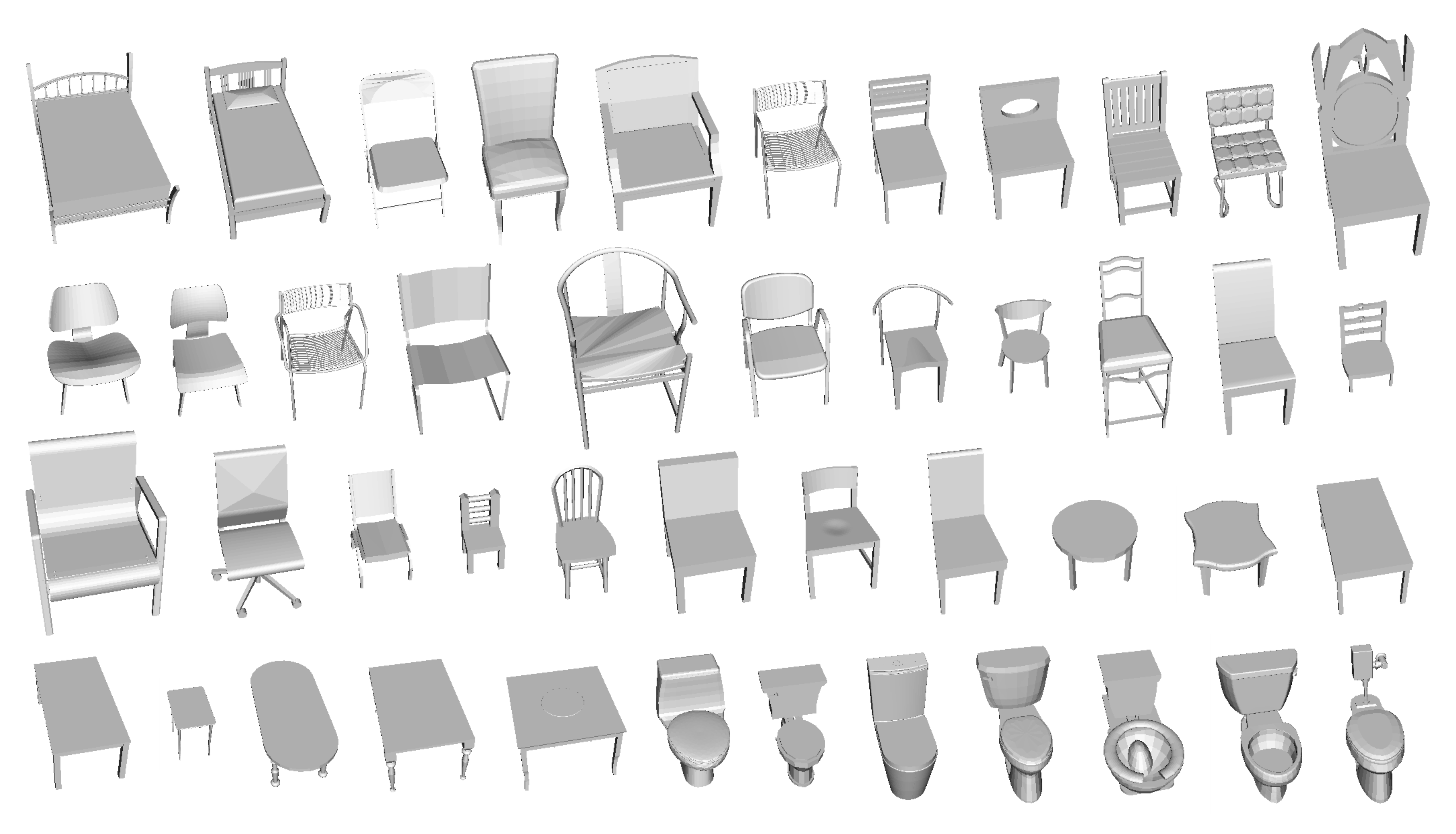}
\vspace{-2em}
\caption{Instances used to train a separate ISA for each category. These training instances are used to generate templates for testing Linemod.}
\vspace{-1.5em}
\label{fig4}
\end{figure*}

\begin{figure*}[!t]
\captionsetup[subfigure]{labelformat=empty}
\centering
\includegraphics[height=1.05in]{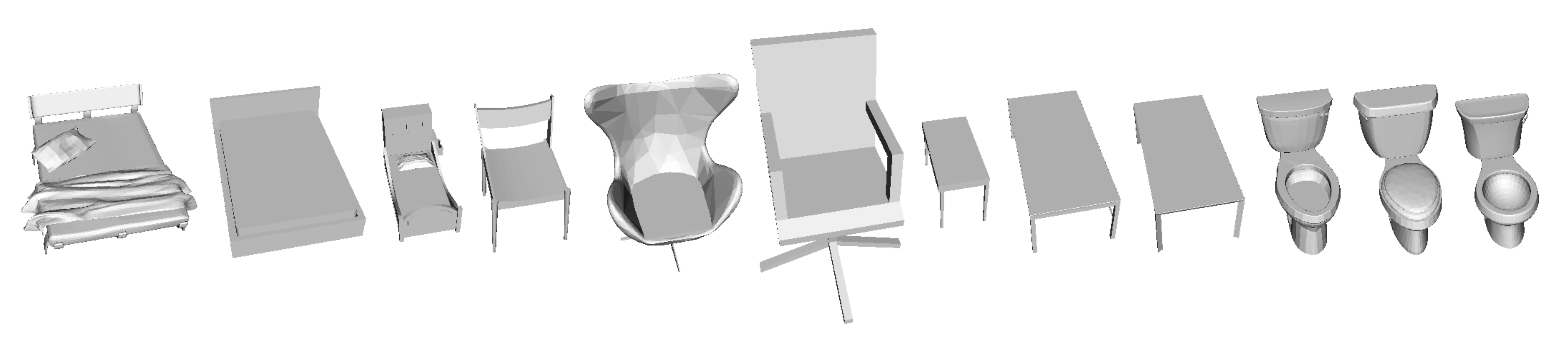}
\caption{Unseen object instances on which ISA and Linemod are tested}
\label{fig5}
\vspace{-2em}
\end{figure*}

\begin{figure*}[!t]
\captionsetup[subfigure]{labelformat=empty}
\centering
\includegraphics[height=2.1in]{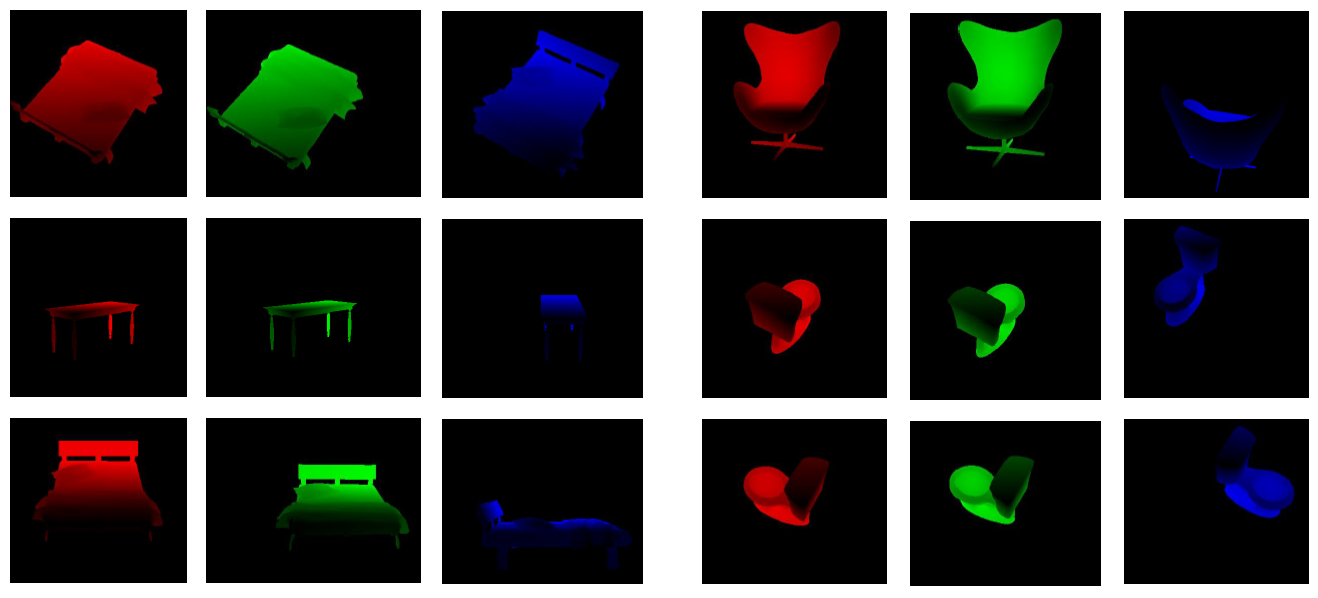}
\caption{Sample results generated by ISA on synthetic data: (for each triplet) each row is of per viewpoint, red is ground truth, green is estimation based on the quality function $Q_1 \&Q_2 \&Q_3$, blue is estimation based on the quality function $Q_1$ only.}
\label{fig6}
\vspace{-1.0em}
\end{figure*}

The main reason why we conduct experiments first on synthetic data is to demonstrate the intrinsic structure adaptation performance of the proposed algorithm in order to have a better understanding on its behaviour across unseen instances.\\
\noindent \textbf{Training ISA.} We employ one-class privileged training using only positive synthetic samples and train the classifiers based on parts extracted from the depth images of the instances in the given categories. Note that, the data related to skeletal representation is only available during training, and in the test phase, the parts reach the leaf nodes using depth appearances in order to vote for a 6D pose. The models from which the depth images are synthesised are sorted through the trainining part of ModelNet10. The number of the instances, the number of the viewpoints from which synthetic depth images are rendered, and the number of the parts used during training are shown in Table \ref{table_analysis}.\\ 
\indent We train $16$ different forests each $4$ of which are individually trained using the quality functions $Q_1$, $Q_1 \& Q_2$, $Q_1 \& Q_3$, and $Q_1 \& Q_2 \& Q_3$ per category. The instances used to train the forests are shown in Fig. \ref{fig4}.\\
\noindent \textbf{Linemod Templates.} Since Linemod is an instance-based detector, the templates method uses are of the object instance on which the the method is tested. Hence, in order to fairly compare Linemod detector with ISA, we employ the following strategy: on the test images of a given category (\textit{e.g.} chair), we run the Linemod detector using the templates of each training instance (for chair, we run Linemod detector $28$ times using $89$ templates of each of $28$ training instances, see Table \ref{table_analysis}). We sort the recall values, and report $3$ different numbers: Linemod (min) represents the lowest recall obtained by any of the training instances, Linemod (max) depicts the highest recall obtained by any of the training instances, and Linemod (all) shows the mean of recall values obtained by all of the training instances.\\
\noindent \textbf{Test.} Unseen test instances are shown in Fig. \ref{fig5}. The resultant recall values are depicted in Table \ref{tab:result} (left). A short analysis on the table reveals that the ISAs based on the 6D pose entropy $Q_1$ demonstrate the poorest performance. Thanks to the utilization of the skeletal link angle entropy $Q_2$, in addition to the 6D quality function, the classifiers reach higher recall values. In case the skeletal node offset entropy $Q_3$ is used along with the 6D pose entropy, there is a relative improvement if we compare with the classifiers trained only on 6D pose entropy. The combined utilization of 6D pose, skeletal link angle, and skeletal node offset entropies performs best on average.\\
\begin{table}[!t]
\caption{Numbers on training samples}
\centering
\begin{tabular}[t]{|c|c|c|c|c|}
\hline
                   &bed         &chair     &table       &toilet      \\
\hline
\#instances        &2           &28        &8           &7           \\
\hline
\#view (per inst.) &89          &89        &89          &89          \\
\hline
\#parts (total)    &$\sim$600k  &$\sim$1m  &$\sim$900k  &$\sim$800k  \\
\hline
\end{tabular}
\label{table_analysis}
\end{table}
\begin{table*}[t]
\caption{\label{tab:result}(left) Comparison on 6D object pose using the evaluation metric in Subsect. \ref{Pose_Definition}. (right) Comparison on 3D object detection using the evaluation metric in \cite{2}.}
\centering
\setlength\tabcolsep{4pt}
\begin{minipage}{0.35\textwidth}
\centering
\resizebox{1.03\columnwidth}{!}{
  \begin{tabular}{ l c c c c c c}
    \toprule
    \textbf{Method}                 &bed           &chair         &table         &toilet           &average \\
    \midrule
    ISA ($Q_1$)                     &39            &40            &50            &80               &52.25\\
    ISA ($Q_1$ \& $Q_2$)            &41            &37            &53            &89               &55.0\\
    ISA ($Q_1$ \& $Q_3$)            &39            &46            &53            &82               &55.0\\
    ISA ($Q_1$ \& $Q_2$ \& $Q_3$)   &46            &42            &52            &87               &56.75\\
    \midrule
    Linemod (min)                   &58            &5             &9             &27               &25\\
    Linemod (max)                   &62            &51            &69            &83               &66\\
    Linemod (all)                   &60            &32            &37            &58               &47\\
    \bottomrule
  \end{tabular}
}
\end{minipage}%
\begin{minipage}{0.63\textwidth}
\centering
\resizebox{0.96\columnwidth}{!}{
    \begin{tabular}{ l c c c c c c}
    \toprule
    \textbf{Method} &input channel &bed           &chair         &table         &toilet &mean     \\
    \midrule    
    Sliding Shapes \cite{2}                       &depth         &33.5          &29            &34.5          &67.3   &41.075    \\
    \cite{4} on instance seg.           &depth         &71            &18.2          &30.4          &63.4   &45.75    \\
    \cite{4} on estimated model         &depth         &72.7          &47.5          &40.6          &72.7   &58.375    \\
    Deep Sliding Shapes \cite{1}                 &depth         &83.0          &58.8          &68.6          &79.2   &72.40    \\
    \midrule
    ISA based on $Q_1 \& Q_2 \& Q_3$    &depth         &52.0          &36.0          &46.5          &67.7   &50.55   \\
    \bottomrule

  \end{tabular}
  }
\end{minipage}%
\end{table*}
\indent For the \textit{bed} category, separately using $Q_2$ and $Q_3$ along with $Q_1$ demonstrates approximately the same performance when the classifiers are trained only on the quality function $Q_1$. However, the combined utilization of $Q_1$, $Q_2$, and $Q_3$ shows the best performance. For the \textit{chair} category, the forest trained on $Q_1$ and $Q_3$ generates the highest recall value, describing the positive impact of using the skeleton node offset entropy. Unlike the bed category, exploiting the skeletal link angle entropy $Q_2$ along with $Q_1 \& Q_3$ relatively degrades the performance of ISA. For the \textit{table} category, one can observe that the skeletal link angle entropy and the skeletal node offset entropy contribute the same to the classifiers in order to generalize across unseen instances. Training the forests using both $Q_1 \& Q_2$ and $Q_1 \& Q_3$ gives rise $3 \%$ improvement with respect to the quality function $Q_1$ only. For the \textit{toilet} category, using $Q_1$ along with $Q_2$ outperforms other forests. Despite the fact that adding the last term $Q_3$ into the combined quality function relatively decreases the recall value, the resultant performance is still better that the classifier trained $Q_1$ only. Figure \ref{fig6} depicts sample hypotheses of unseen instances with ground truth poses in red. The forests based on $Q_1 \& Q_2 \& Q_3$ hypothesise the green estimations which are considered as true positive, and the forests based on $Q_1$ hypothesise the blue estimations which are considered as false positive. Note that, both 3D position and 3D orientation of an esimation are used when deciding whether the object is correctly estimated.\\
\indent In Table \ref{tab:result} (left), we report recall values for the Linemod detector. Using the templates of each training instance of the given category, we run Linemod, and sort the recall values. According to the Linemod (min) recall values, Linemod worst performs on the \textit{chair} category, whilst it shows best performance on the \textit{toilet} category. The maximum recall value that Linemod achieve is of the \textit{toilet} category. When we compute the mean for all recall values, Linemod best performs on the \textit{bed} category.
\vspace{-1.0em}
\subsection{Experiments on Real Data}
\begin{figure}[!t]
\captionsetup[subfigure]{labelformat=empty}
\centering
\includegraphics[height=2.4in]{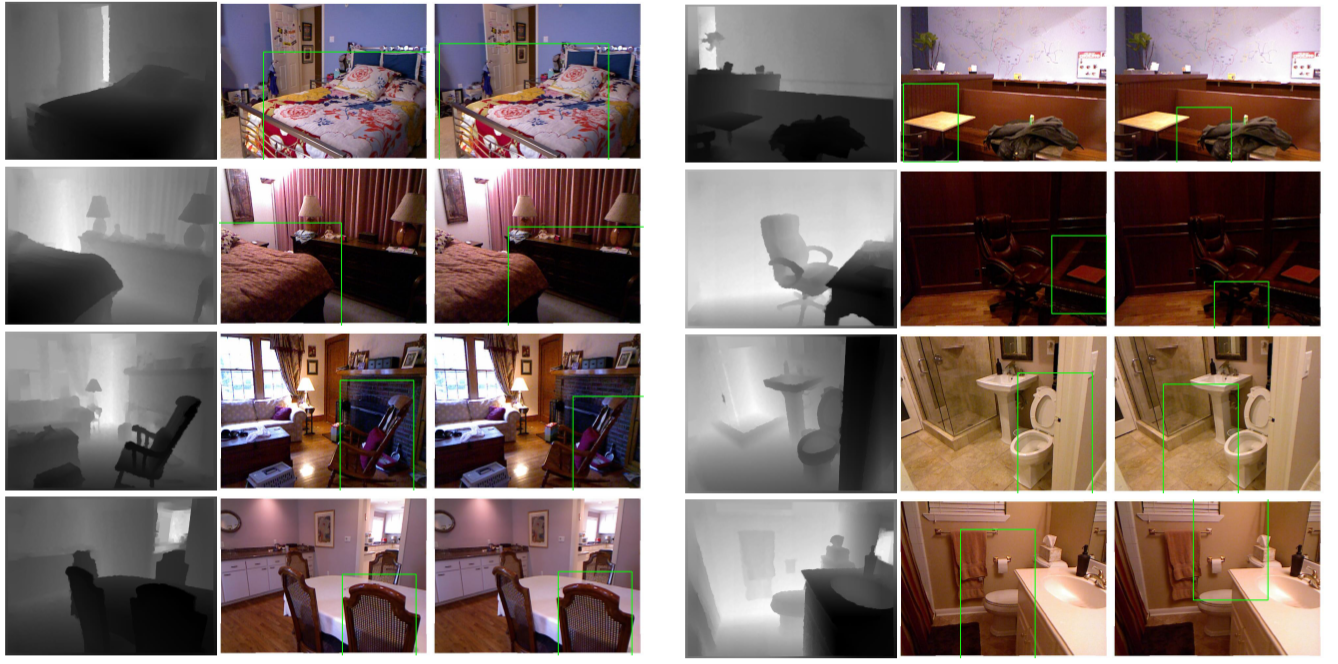}
\caption{Sample results generated by ISA on real data: (for each triplet) each row is for per scene. First column depicts depth images of scenes. Estimations in the middle belong to ISAs trained using $Q_1\&Q_2\&Q_3$, and hypotheses on the right are of ISAs trained on $Q_1$ only.}
\label{fig7_8}
\vspace{-2.0em}
\end{figure}

\indent Table \ref{tab:result} (right) depicts the comparison on 3D object detection. A short analysis on the table reveals that our architecture demonstrate $50 \%$ average precision. The highest value ISA reaches is on the \textit{toilet} category, mainly because of the limited deviation in shape in between the instances. ISA next best performs on \textit{bed}, with $52 \%$ mean precision. The accuracy on both the categories \textit{bed} and \textit{table} are approximately the same. Despite the fact that all forests used in the experiments undergo relatively a naive training process, the highest number of the instances during training are used for the chair category. However, ISA worst performs on this category, since the images in the test dataset have strong challenges of the instances, such as occlusion, clutter, and high diversity from the shape point of view. We lastly present sample results in Fig. \ref{fig7_8}. In these figures, the leftmost images are the inputs of our architecture, and the $2^{nd}$ and the $3^{rd}$ columns demonstrate the estimations of the forests based on $Q_1\&Q_2\&Q_3$ and $Q_1$ only, respectively.\\
\vspace{-2.5em}
\section{Conclusion}
\vspace{-1.0em}
In this paper we have introduced a novel architecture, ISA, for category-level 6D object pose estimation from depth images. We have designed the proposed architecture in such a way that the challenges of the categories, intra-class variations, distribution shifts among source and target domains, can successfully be tackled while the 6D pose of unseen objects are estimated. To this end, we have engineered ISA based on the following technologies: We have firstly presented Semantically Selected Centers (SSC) for the category-level 6D object pose estimation problem. We next have utilized 3D skeleton structures and derived those as shape-invariant features. Using these features, we have represented the parts extracted from the instances of given categories, and employed privileged one-class learning based on these parts. We have performed graph matching during training so that the adaptation capability of the proposed architecture is improved across unseen instances. Experiments conducted on test images validate the promising performance of ISA. In the future, we are planning to improve the performance of ISA approaching the problem from transfer learning point of view.
\clearpage


\end{document}